# Multi-Label Classification Method Based on Extreme Learning Machines

Rajasekar Venkatesan
School of Electrical and Electronics Engineering
Nanyang Technological Institute
Singapore
RAJA0046@e.ntu.edu.sg

Meng Joo Er
School of Electrical and Electronic Engineering
Nanyang Technological Institute
Singapore
EMJER@ntu.edu.sg

*Abstract*— **In this paper, an Extreme Learning Machine (ELM) based technique for Multi-label classification problems is proposed and discussed. In multi-label classification, each of the input data samples belongs to one or more than one class labels. The traditional binary and multi-class classification problems are the subset of the multi-label problem with the number of labels corresponding to each sample limited to one. The proposed ELM based multi-label classification technique is evaluated with six different benchmark multi-label datasets from different domains such as multimedia, text and biology. A detailed comparison of the results is made by comparing the proposed method with the results from nine state of the arts techniques for five different evaluation metrics. The nine methods are chosen from different categories of multi-label methods. The comparative results shows that the proposed Extreme Learning Machine based multi-label classification technique is a better alternative than the existing state of the art methods for multi-label problems.**

## I. INTRODUCTION

In general, classification in machine learning corresponds to assignment of a single target label for the input sample instances. As only one label from a set of disjoint labels is assigned to the input data, this type of classification is called single label classification. However, there are several conditions where the input data falls under more than one class. This condition of classification, where the input data correspond to a set of class labels instead of one, is called Multi-label classification. Initially, the application of multi-label classification is primarily focused on text-categorization [1-5] and medical diagnosis [6]. But recent realization of the omnipresence of multi-label prediction tasks in real world problems drawn more and more research attention to this domain [7]. The application of multi-label classification has extended to other areas such as bioinformatics [8-9], scene classification [10-11], map labelling [12] etc.

Single label classification is a common learning problem where each instance is associated with a unique class label from a set of disjoint labels L. Unlike single label classification, multi-label classification enables each instance to be associated with more than one class. That is, in multi-label classification, each instance belongs to a subset of classes from L. Thus, binary classification, multi-class classification and ordinal regression problems can be seen as special cases of multi-label problems where the number of labels assigned to each instance is equal to 1 [13].

In recent years, several techniques are developed and is available in the literature that are used to perform multi-label classification. Gjorgi et al. in their paper [14] categorizes these techniques into three major categories. An extensive comparison of multi-label methods has been performed by Gjorgi et al. [14] and from the comparison of results it can be seen that there exists no single method that performs uniformly well on a wide range of datasets. Each method outperforms other in only a few datasets and performs less efficiently in other datasets. In this article, we propose an Extreme Learning Machine (ELM) based multi-label learning method which performs effectively in a wide range of datasets. The proposed method outperforms all or most of existing methods in different performance metrics. ELM based multi-label learning is never implemented in the literature thus far.

The rest of this article is organized as follows. Section II gives a brief overview of multi-label classification problem and various methods used. Section III presents the proposed algorithm and Section IV discusses the different benchmark metrics for dataset specification and algorithm evaluation. Section V provides the experimentation specifications and Section VI discusses the results comparison with existing methods and related discussions. Finally, in Section VII, the contribution of the article is summarized and concluded.

## II. MULTI-LABEL LEARNING

As given by Andre et al. [15], the term "classification" can be formally defined as, *"Given a set of training examples composed of pairs {$x_i$, $y_i$}, find a function f(x) that maps each attribute vector xi to its associated class $y_i$, i = 1,2,3…, n, where n is the total number of training examples."* Single label classification involves associating a single label 'l' from a set of disjoint labels 'L' to each of the input data sequence [16]. There are two sub categories of single label classification. They are binary classification and multi-class classification. Binary classification (L=2) involves classifying the input data samples into either of two sets based on a specific classification metric. Disease diagnosis, quality control are some of the major application areas of this method. On the other hand, Multi-class classification (L>2) involves classifying the input samples into more than two classes. There are several multi-class data sets such as iris, waveform, balance scale, glass, dna etc.

In contrast to single label classification, in multi-label classification, each of the input samples belongs to more than one of the classification labels. For each input sample, there exists a set of labels M, which is a subset of L to which the input sample belongs to. The application areas of multi-label classification is expanding in recent years. Traditional binary and multi-class classification problems forms the special class of multi-label classification. But the generality of multi-label classification makes it more difficult to be implemented and trained than the others [17]. Multi-label classification has applications in various domains such as text categorization, protein function classification, music categorization, semantic scene classification and several upcoming new domains. Several multi-label classification techniques has been developed and are currently available in the literature. The paper [14] discuss in detail the state of the arts multi-label classification methods and categorizes the existing methods into three groups. Adapted from [14], the overview of existing methods can be summarized as shown in figure 1.

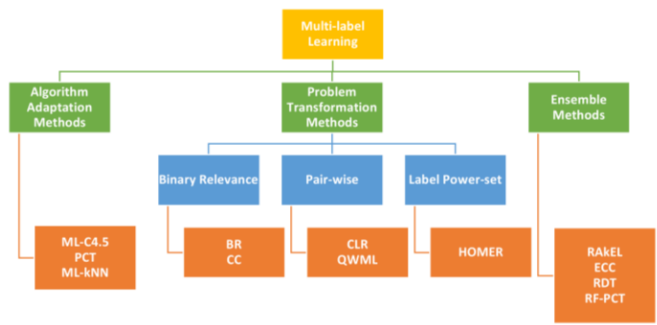

Figure 1. Overview of existing methods

The initial overview of the multi-label classification methods is presented in [16], which classifies the existing methods then into two categories. Algorithm Adaptation (AA) methods and Problem Transformation (PT) methods. In recent years, more multi-label classifiers have been developed and the most recent overview of multi-label classifiers introduces the third category of Ensemble (EN) methods [14].

*A. Algorithm Adaptation Methods*

The AA methods are those that can adapt, extend and customize an existing machine learning algorithm to meet the needs of solving multi-label problems [14]. Hence the AA methods can be subcategorized based on which of the existing algorithm, the multi-label variant is developed. Multi-label variants are developed based on Boosting, kNN, Decision Trees, Neural Networks, and SVM [14 and references within].

*B. Problem Transformation Methods*

The PT methods employ a unique transformation that converts the multi-label problem into one or more single-label problems. Some of the early PT methods use simple transformation techniques like instance elimination, label elimination, label decomposition etc. More advanced transformation methods like copy transformation, dubbed copy transformation, Label powerset, pruned problem transformation methods are developed subsequently. More popular and novel transformation methods such as Binary relevance (BR) method and Classifier Chain (CC) method are developed in recent years. These methods decompose the multi-label classification into series of single-label classification problem with each single label problem focusing on one label of the multi-label case. The Label powerset methods like HOMER combines the multiple labels and creates new labels thus making it into single-label problem. Pair-wise methods use multiple classifiers that cover all possible label pairs. To combine the output of the classifiers either voting based method is used as in CLR or Qweighted approach is used as in QWML.

*C. Ensemble Methods*

The EN method based multi-label classifiers use an ensemble of AA or PT classifiers to address the multi-label classification problem. Some of the widely known ensemble methods are Random k labelsets (RAkEL), Ensemble of Classifier Chains (ECC), Random forest based predictive clustering trees (RF-PCT), Random Decision Tree (RDT) etc. Based on the machine learning algorithm used, the multi-label methods have been grouped as shown in figure 2 [14].

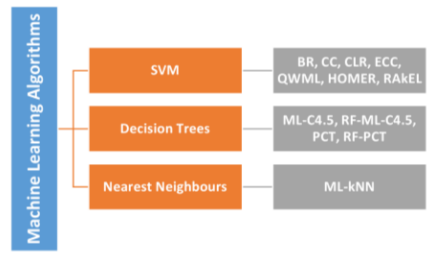

Figure 2. Multi-label methods based on machine learning algorithm

It can be seen from the brief review that ELM based techniques are thus far not used to implement the multi-label classification problem. This paper proposes an AA method multi-label classifier based on ELM.

### III. PROPOSED ALGORITHM

This section provides a brief review of the ELM technique and the proposed algorithm used for multi-label classification.

Consider there are N training samples represented as {($x_i$, $y_i$)} where i varies from 1 to N, $x_i$ denotes the input data vector and $y_i$ denotes the output. Let there be N' number of hidden neurons in the network, the output of the basic batch learning ELM technique can be given as in (1).

$$f_{N'}(x) = \sum_{i=1}^{N'} \beta_i G(w_i, b_i, x), \quad x \in R^n, w_i \in R^n, \beta_i \in R^m \quad (1)$$

From the theory of ELM it is evident that the input weights ($w_i$) and the hidden layer neuron bias ($b_i$) can be randomly assigned. Thus, the output weight is determined as $\beta = H^+Y$ where $H^+$ is the Moore-Penrose generalized inverse of hidden layer output matrix and $Y = [y_1,...,y_N]^T$.

*A. Initialization of Parameters*

Basic parameters such as the type of activation function and number of hidden layer neurons in the ELM are initialized.

*B. Processing of Inputs*

The multi-label output corresponding to the input sequence is in general provided as '0' and '1' for each label, with '1' corresponding to the labels to which the input samples belong to. For multi-label classification, each of the input samples may belong to one or more than one label. These inputs are converted from unipolar to bipolar representation.

*C. ELM Training*

The input weights ($w_i$) and the bias of the hidden layer neurons ($b_i$) are randomly assigned. The basic batch learning equation as given in (1) can be compactly written as, **$H\beta = Y$,** where H is an N x N' matrix, β is N' * L matrix and Y is N x L output matrix, where each row gives the output of corresponding input sample and each column corresponds to each of the labels. In the training phase, the output weight, β is evaluated from the training input and output data as, **$\beta = H^+Y$,** where $H^+$ is the Moore-Penrose generalized inverse of hidden layer output matrix and Y is the bipolar N x L matrix.

*D. ELM Testing*

In the testing phase, the test output is evaluated by using the β values found during the training phase in the equation Y = Hβ. In single label classification, the class label to which the input sample belong to can be identified by determining the column which has the maximum value in Y. The label corresponding to the column with maximum value is identified as the classifier output. In contrast, in multi-label classification problems, the input samples may belong to one or more labels and hence it cannot be directly identified by identifying the column with maximum value. For multi-label classification, The N x L values of Y are passed as arguments to the bipolar step function. A threshold of '0' is applied to the resultant values. The set of columns of the resulting matrix with values 1 gives the multi-label belongingness of the corresponding input.

## IV. MULTI-LABEL CLASSIFICATION METRICS

*A. Dataset Metrics*

The degree of multi-label nature of each data set varies. Some data sets may have large part of the input data set to be multi-label in nature whereas some other data sets may have only a few multi-label samples. The degree of multi-label nature can be quantified using two metrics: Label cardinality ($L_c$) and Label density ($L_d$). Let the dataset be given as {xi, Yi}, i = 1...N with L number of labels. Then, Label Cardinality is defined as the average number of labels of the input samples in the dataset. Label density is given as the average number of labels of the input samples divided by number of labels.

$$Label - cardinality = \frac{1}{N}\sum_{i=1}^{N}|Y_i| \quad (2)$$

$$Label - density = \frac{1}{N}\sum_{i=1}^{N}\frac{|Y_i|}{L} \quad (3)$$

Label cardinality signifies the average number of labels present in the training data set. Label cardinality is independent of the number of labels present in the dataset. Label density takes into consideration the number of labels present in the dataset. Two datasets having same label cardinality, but different label density can differ largely in their properties and may cause different behavior to the training algorithm [17].

*B. Evaluation Metrics*

Multi class classification is unique in its nature from single label classification in which the classification can be partially correct. In single label classification problems, the classification can only be either correct or wrong. Partial correctness are not observed in single label classification. Whereas in multi-label classification problems, the classifier may classify at least one of the classes correctly and one or more of the classes in a wrong manner thus resulting in the partial correctness of the classification. Thus the traditional performance evaluation metrics used for single label classification cannot be used to evaluate the performance of the multi-label classification case. Multi-label classification requires a different set of performance metrics to evaluate the effectiveness and efficiency of the training method. The performance metrics are proposed in the literature used to validate the multi-label learning. They are Hamming Loss, Accuracy, Precision, Recall and F1-measure. There are few other evaluation measures like one-error, coverage etc. which are used for ranking based multi-label classification methods.

## V. EXPERIMENTATION

In this section, we present the experimental design used to evaluate the proposed method and compare its performance with other existing methods.

The proposed method is tested for its performance using six different standard benchmark datasets. The datasets are chosen from different domains such as Multimedia, Text and Biology. A diverse nature of datasets are chosen so as to verify for the consistency, robustness and reliability of the proposed method for generic environment. Certain existing techniques performs well in a specific datasets but their performance reduces significantly when introduced with different datasets. Hence consistent performance of the method is of critical importance when to be used real-time, real world applications. The number of labels varies from as low as 6 labels to as high as 374 labels. Datasets with number of attributes varying from 72 to as high as 1449 attributes are tested and verified.

Multi-label datasets has another unique nature that each of the datasets are not equally multi-labelled. The multi-label nature of the datasets varies. The degree of multi-labelness is quantified by the use of metrics label cardinality and label density. Label cardinality gives the average number of labels of the input samples in the dataset. For single-label classification, the label cardinality is always 1. But for multi-label problems, the input samples can have one or more than one label associated with them. Hence the label cardinality for multi-label problems is always greater than 1. The datasets used for experimental validation of the proposed method has its label cardinality varying from 1.07 to 4.24. Label cardinality of 4.24 signifies that each sample has an average of more than 4 labels associated with it. Label density takes into account the number of labels along with the average number of labels of the input samples. The lower the label density the lesser the number of occurrence of the label in the dataset. Lower label density indicates that there are only fewer samples corresponding to each label and hence the learning method needs to learn the label within those limited samples. Thus the label density of 0.009 indicates that the average percentage of occurrence of each label in dataset is only 0.9%. The datasets are taken from the KEEL multi-label dataset repository. The specifications of the datasets used for validation is given in table 1.

TABLE I. SPECIFICATIONS OF DATASETS

| Dataset | Domain | #Attr | L | #train | #test | $L_c$ | $L_d$ |
|---|---|---|---|---|---|---|---|
| Emotions | Multimedia | 72 | 6 | 400 | 193 | 1.87 | 0.312 |
| Yeast | Biology | 103 | 14 | 1600 | 817 | 4.24 | 0.303 |
| Scene | Multimedia | 294 | 6 | 2000 | 407 | 1.07 | 0.178 |
| Corel5k | Multimedia | 499 | 374 | 4500 | 500 | 3.52 | 0.009 |
| Enron | Text | 1001 | 53 | 1200 | 502 | 3.38 | 0.064 |
| Medical | Text | 1449 | 45 | 700 | 278 | 1.25 | 0.027 |

TABLE II. SPECIFICATIONS OF METHODS USED FOR COMPARISON

| Method Name | Method Category | Machine Learning Category |
|---|---|---|
| Classifier Chain (CC) | PT | SVM |
| QWeighted approach for Multi-label Learning (QWML) | PT | SVM |
| Hierarchy Of Multi-label ClassifiERs (HOMER) | PT | SVM |
| Multi-Label C4.5 (ML-C4.5) | AA | Decision Trees |
| Predictive Clustering Trees (PCT) | AA | Decision Trees |
| Multi-Label k-Nearest Neighbors (ML-kNN) | AA | Nearest Neighbors |
| Ensemble of Classifier Chains (ECC) | EN | SVM |
| Random Forest Predictive Clustering Trees (RF-PCT) | EN | Decision Trees |
| Random Forest of ML-C4.5 (RFML-C4.5) | EN | Decision Trees |

The proposed method is validated with the datasets given in table. The results achieved by the proposed method is compared with the state of the arts techniques. State of the arts techniques from different categories of multi-label classification such as AA methods, PT methods, EN methods are used for result comparison. Three methods from each of the PT, AA and EN methods are used for result comparison. The methods from different machine learning techniques like SVM, Decision Trees and Nearest Neighbors are used for comparison. The details of the state of the art methods used for result comparison with proposed method is given in table 2. The proposed method is verified with 9 state of the art methods and 6 benchmark datasets. Hamming Loss, Accuracy, Precision, Recall and F-measure are used as evaluation metrics to compare the proposed method with existing algorithms.

## VI. RESULTS AND DISCUSSION

The five evaluation metrics are evaluated for each of the datasets using the proposed method. The results obtained are discussed in the order of the evaluation metrics.

Consider the dataset be given as $\{x_i, Y_i\}$, $i = 1\ldots N$ with L number of labels. Let MLC be the training method and $Z_i = MLC(x_i)$ be the output labels predicted by the classification method. The expression to identify the evaluation metrics are given in equations.

$$Hamming\_Loss = \frac{1}{N}\sum_{i=1}^{N}\frac{1}{L}|MLC(x_i \Delta Y_i)| \tag{4}$$

$$Accuracy = \frac{1}{N}\sum_{i=1}^{N}\left(\frac{|MLC(x_i) \cap Y_i|}{|MLC(x_i) \cup Y_i|}\right) \tag{5}$$

$$Precision = \frac{1}{N}\sum_{i=1}^{N}\left(\frac{|MLC(x_i) \cap Y_i|}{|MLC(x_i)|}\right) \tag{6}$$

$$Recall = \frac{1}{N}\sum_{i=1}^{N}\left(\frac{|MLC(x_i) \cap Y_i|}{|Y_i|}\right) \tag{7}$$

$$F1-measure = \frac{1}{N}\sum_{i=1}^{N}\left(\frac{2*|MLC(x_i) \cap Y_i|}{|MLC(x_i)|+|Y_i|}\right) \tag{8}$$

### A. Hamming Loss

Hamming loss gives the percentage of wrong labels to the total number of labels. Lower the hamming loss, better is the performance of the method used. For an ideal classifier, hamming loss is 0. The hamming loss can be calculated by (4).

### B. Accuracy

Accuracy of the multi-label classifier is defined as the proportion of the predicted correct labels to the total number of labels for that instance. Overall accuracy is the average across all instances. The Accuracy can be evaluated using (5).

### C. Precision

Precision is the proportion of the predicted correct labels to the total number of actual labels averaged over all instances. In other words, it is the ratio of true positives to the sum of true positives and false positives averaged over all instances. The expression for precision is given in (6).

### D. Recall

Recall is the proportion of the predicted correct labels to the total number of predicted labels averaged over all instances. In other words, it is the ratio of true positives to the sum of true positives and false negatives averaged over all instances. The expression for recall is given in (7).

### E. F1-measure

F1 measure is given by the harmonic mean of Precision and Recall. The expression to evaluate F1 measure is given in (8).

Higher the values of accuracy, precision, recall and F1-measure, better the performance of the proposed method. And lower the hamming loss corresponds to better accuracy of the proposed method. Comparison of the hamming loss, accuracy, precision, recall and F1-measure obtained by the proposed method and other existing methods is shown in tables 3-7. The results of the state of the art methods are obtained from [14]. The values of the evaluation metrics which are equal to or greater than the value obtained by the proposed method is highlighted in blue. It is evident from the table that the proposed method is consistently better than most of the existing methods in all the five evaluation metrics. The performance of the methods for the datasets used are also shown in figures 3-7.

From figure 3 we can see that, the proposed ELM based method lower hamming loss than the existing methods and is ranked first in yeast, corel5k and medical datasets and is almost as nearly as the first in Enron dataset. Though it is not the first ranked in hamming loss performance for emotion and scene recognition, it still is one of the top classifiers when compared to other methods. Also, it can be seen from the figure that, no existing algorithm is consistent in their performance throughout all the datasets. Also from the figures 3 to 7, it is clear that the proposed method consistently gives better performance than the existing methods throughout all the datasets across all the evaluation metrics observed. Some of the methods outperform the proposed method in one or two datasets, but loses its performance for other datasets.

TABLE III. COMPARISON OF HAMMING LOSS

| Dataset | CC | QWML | HOMER | ML-C4.5 | PCT | ML-kNN | ECC | RFML-C4.5 | RF-PCT | ELM |
|---|---|---|---|---|---|---|---|---|---|---|
| Emotion | 0.256 | 0.254 | 0.361 | 0.247 | 0.267 | 0.294 | 0.281 | 0.198 | 0.189 | **0.251** |
| Yeast | 0.193 | 0.191 | 0.207 | 0.234 | 0.219 | 0.198 | 0.207 | 0.205 | 0.197 | **0.191** |
| Scene | 0.082 | 0.081 | 0.082 | 0.141 | 0.129 | 0.099 | 0.085 | 0.116 | 0.094 | **0.085** |
| Corel5k | 0.017 | 0.012 | 0.012 | 0.010 | 0.009 | 0.009 | 0.009 | 0.009 | 0.009 | **0.009** |
| Enron | 0.064 | 0.048 | 0.051 | 0.053 | 0.058 | 0.051 | 0.049 | 0.047 | 0.046 | **0.047** |
| Medical | 0.077 | 0.012 | 0.012 | 0.013 | 0.023 | 0.017 | 0.014 | 0.022 | 0.014 | **0.011** |

TABLE IV. COMPARISON OF ACCURACY

| Dataset | CC | QWML | HOMER | ML-C4.5 | PCT | ML-kNN | ECC | RFML-C4.5 | RF-PCT | ELM |
|---|---|---|---|---|---|---|---|---|---|---|
| Emotion | 0.356 | 0.373 | 0.471 | 0.536 | 0.448 | 0.319 | 0.432 | 0.488 | 0.519 | **0.412** |
| Yeast | 0.527 | 0.523 | 0.559 | 0.480 | 0.440 | 0.492 | 0.546 | 0.453 | 0.478 | **0.514** |
| Scene | 0.723 | 0.683 | 0.717 | 0.569 | 0.538 | 0.629 | 0.735 | 0.388 | 0.541 | **0.676** |
| Corel5k | 0.030 | 0.195 | 0.179 | 0.002 | 0.000 | 0.014 | 0.001 | 0.005 | 0.009 | **0.044** |
| Enron | 0.334 | 0.388 | 0.478 | 0.418 | 0.196 | 0.319 | 0.462 | 0.374 | 0.416 | **0.418** |
| Medical | 0.211 | 0.658 | 0.713 | 0.730 | 0.228 | 0.528 | 0.611 | 0.250 | 0.591 | **0.715** |

TABLE V. COMPARISON OF PRECISION

| Dataset | CC | QWML | HOMER | ML-C4.5 | PCT | ML-kNN | ECC | RFML-C4.5 | RF-PCT | ELM |
|---|---|---|---|---|---|---|---|---|---|---|
| Emotion | 0.551 | 0.548 | 0.509 | 0.606 | 0.577 | 0.502 | 0.580 | 0.625 | 0.644 | **0.548** |
| Yeast | 0.727 | 0.718 | 0.663 | 0.620 | 0.705 | 0.732 | 0.667 | 0.738 | 0.744 | **0.718** |
| Scene | 0.758 | 0.711 | 0.746 | 0.592 | 0.565 | 0.661 | 0.770 | 0.403 | 0.565 | **0.685** |
| Corel5k | 0.042 | 0.326 | 0.317 | 0.005 | 0.000 | 0.035 | 0.002 | 0.018 | 0.030 | **0.144** |
| Enron | 0.464 | 0.624 | 0.616 | 0.623 | 0.415 | 0.587 | 0.652 | 0.690 | 0.709 | **0.668** |
| Medical | 0.217 | 0.697 | 0.762 | 0.797 | 0.285 | 0.575 | 0.662 | 0.284 | 0.635 | **0.774** |

TABLE VI. COMPARISON OF RECALL

| Dataset | CC | QWML | HOMER | ML-C4.5 | PCT | ML-kNN | ECC | RFML-C4.5 | RF-PCT | ELM |
|---|---|---|---|---|---|---|---|---|---|---|
| Emotion | 0.397 | 0.429 | 0.775 | 0.703 | 0.534 | 0.377 | 0.533 | 0.545 | 0.582 | **0.491** |
| Yeast | 0.600 | 0.600 | 0.714 | 0.608 | 0.490 | 0.549 | 0.673 | 0.491 | 0.523 | **0.608** |
| Scene | 0.726 | 0.709 | 0.744 | 0.582 | 0.539 | 0.655 | 0.771 | 0.388 | 0.541 | **0.709** |
| Corel5k | 0.056 | 0.264 | 0.250 | 0.002 | 0.000 | 0.014 | 0.001 | 0.005 | 0.009 | **0.043** |
| Enron | 0.507 | 0.453 | 0.610 | 0.487 | 0.229 | 0.358 | 0.560 | 0.398 | 0.452 | **0.508** |
| Medical | 0.754 | 0.801 | 0.760 | 0.740 | 0.227 | 0.547 | 0.642 | 0.251 | 0.599 | **0.744** |

TABLE VII. COMPARISON OF F1-MEASURE

| Dataset | CC | QWML | HOMER | ML-C4.5 | PCT | ML-kNN | ECC | RFML-C4.5 | RF-PCT | ELM |
|---|---|---|---|---|---|---|---|---|---|---|
| Emotion | 0.461 | 0.481 | 0.614 | 0.651 | 0.554 | 0.431 | 0.556 | 0.583 | 0.611 | **0.518** |
| Yeast | 0.657 | 0.654 | 0.687 | 0.614 | 0.578 | 0.628 | 0.670 | 0.589 | 0.614 | **0.658** |
| Scene | 0.742 | 0.710 | 0.745 | 0.587 | 0.551 | 0.658 | 0.771 | 0.395 | 0.553 | **0.697** |
| Corel5k | 0.048 | 0.292 | 0.280 | 0.003 | 0.000 | 0.021 | 0.001 | 0.008 | 0.014 | **0.033** |
| Enron | 0.484 | 0.525 | 0.613 | 0.546 | 0.295 | 0.445 | 0.602 | 0.505 | 0.552 | **0.577** |
| Medical | 0.337 | 0.745 | 0.761 | 0.768 | 0.253 | 0.560 | 0.652 | 0.267 | 0.616 | **0.759** |

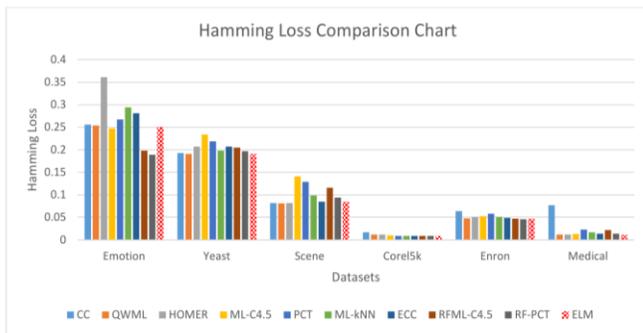

Figure 3. Hamming loss comparison chart

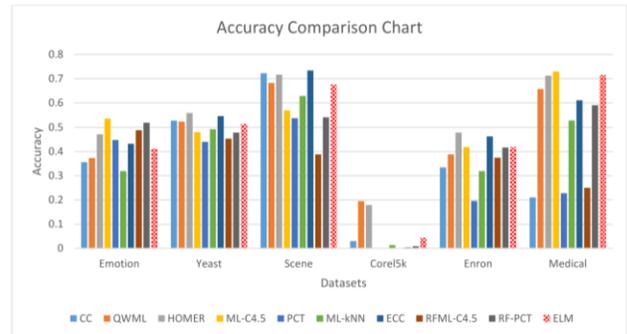

Figure 4. Accuracy comparison chart

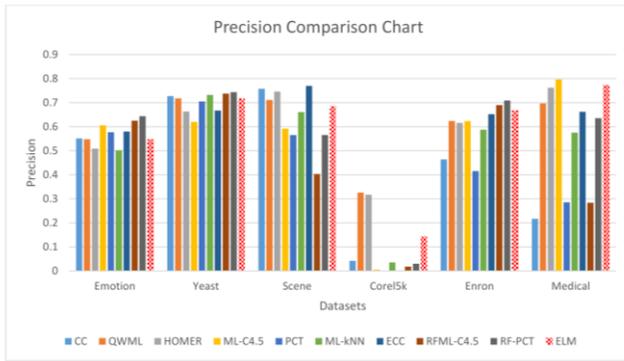

Figure 5. Precision comparison chart

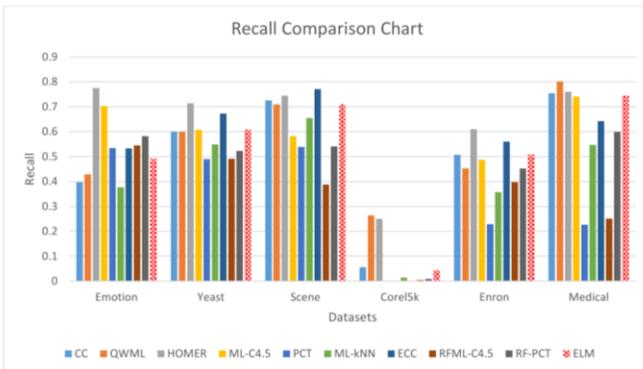

Figure 6. Recall comparison chart

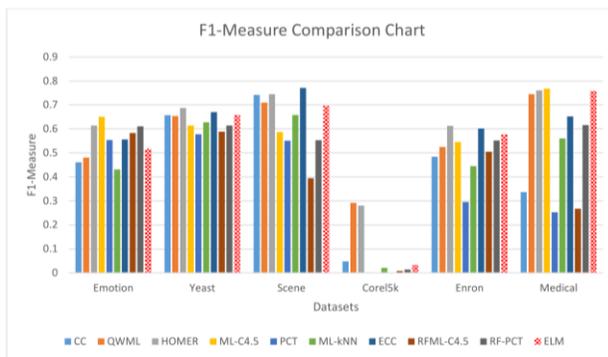

Figure 7. F1-measure comparison chart

The proposed method also exploits the high speed learning which is inherent to the extreme learning machine. The learning speed of ELM is several folds greater than the traditional neural networks. Hence the proposed algorithm can be trained with high speed. Thus the proposed ELM based method will be a better solution for multi-label problems.

VII. CONCLUSION

In this paper, an Extreme Learning Machine based learning technique for Multi-label classification is developed. It is to be noted that ELM based multi-label classification has never been implemented in the literature thus far. In Multi-label classification each of the input samples may belong to one or more than one of the labels. The proposed ELM based method is evaluated using different datasets. The results obtained is compared with several existing state of the art methods. The results show that the proposed method performs effectively than the existing method in most cases and in all the evaluations metrics. Thus the ELM based Multi-label classifier can be a better alternative to solve a wide range of multi-label classification problems from various domains.

ACKNOWLEDGMENT

The first author would like to thank Nanyang Technological University, Singapore for the NTU Research Student Scholarship.